\documentclass[journal]{IEEEtran}
\usepackage{amsfonts}
\usepackage{subfigure}
\usepackage{mathrsfs}
\usepackage{float}
\usepackage{graphics}
\usepackage{epstopdf}
\usepackage{times}
\usepackage{booktabs}
\usepackage{amsmath}
\usepackage{amssymb}
\usepackage{indentfirst,bm}
\usepackage[dvips]{graphicx}
\usepackage{epsfig}
\usepackage{epsf}
\usepackage{color}
\usepackage{cite}
\usepackage{enumerate}
\usepackage{multirow}
\usepackage{stfloats}
\usepackage{amsmath}
\usepackage{tikz}
\usetikzlibrary{positioning, fit, calc, shapes.geometric, arrows.meta}
\usepackage{caption}
\usepackage{algorithm}
\usepackage{algorithmicx}
\usepackage{algpseudocode,lineno}

\usepackage{booktabs}
\usepackage{tabularx}
\usepackage{array}

\usepackage{hyperref}

\usepackage[compact]{titlesec} 
\titlespacing*{\section}{0pt}{*1.3}{*0.8}
\titlespacing*{\subsection}{0pt}{*1.2}{*0.7}

\usepackage{tikz}
\usetikzlibrary{shapes.geometric, arrows}

\tikzstyle{startstop} = [rectangle, rounded corners, minimum width=2.5cm, minimum height=1cm, text centered, draw=black, fill=blue!30]
\tikzstyle{process} = [rectangle, minimum width=2.5cm, minimum height=1cm, text centered, draw=black, fill=orange!30]
\tikzstyle{decision} = [diamond, aspect=2, text centered, draw=black, fill=yellow!30]
\tikzstyle{arrow} = [thick,->,>=stealth]
\usepackage{amsthm}

\newtheoremstyle{mydefstyle}
  {0.5em} 
  {0.5em} 
  {\itshape} 
  {} 
  {\bfseries} 
  {.} 
  { } 
  {} 

\theoremstyle{mydefstyle}

\begin{document}

\title{High-Order Liquid Evidence Modeling for Continuous and Subtle GNSS Spoofing Detection in Autonomous Driving}

\author{Muhammad Ayub Sabir, Junbiao~Pang, Fatima~Ashraf
\IEEEcompsocitemizethanks{

\IEEEcompsocthanksitem Sabir, J. Pang, and Fatima are with the Faculty of Information Technology, Beijing University of Technology, Beijing 100124, China (e-mail: sabir@emails.bjut.edu.cn; junbiao\_pang@bjut.edu.cn;
fatimaashraf@emails.bjut.edu.cn)

 }
}


\maketitle


\begin{abstract}

Continuous and subtle GNSS spoofing poses a serious threat to autonomous vehicles because forged positions may remain locally plausible while gradually becoming inconsistent with vehicle motion observed by non-GNSS onboard sensors. Existing AV-oriented detectors commonly rely on residual thresholds or feature-level classification and provide limited modeling of how weak GNSS--motion inconsistency develops and persists over time. This paper formulates subtle GNSS spoofing detection as a causal sequential evidence-modeling problem and proposes a high-order liquid evidence detector. The method first compares the displacement implied by consecutive GNSS positions with that inferred from independent onboard motion observations and converts their difference into uncertainty-normalized residual evidence. It then represents the current inconsistency, its local evolution, excess above the normal level, accumulated persistence, and displacement validity as causal weak evidence. These cues are mapped into instantaneous, evolutionary, and persistent latent states, aligned through a bounded Kirchhoff-inspired symmetric exchange, and combined through an explicit third-order interaction to capture their coordinated support for spoofing. To model how this coordinated evidence develops over time, second-order liquid dynamics track its memory and evolution to estimate causal spoofing probabilities, which are converted into confirmed alarms using validation-selected threshold and persistence parameters. Experiments on the AV--GPS dataset family demonstrate strong controlled and external generalization, together with clear sequential alarm behavior. On Dataset-1, the proposed detector achieves an AUROC of 0.9932 and an AUPRC of 0.9843, while obtaining the lowest false-positive rate among the learning-based baselines. The source code is publicly available at
\url{https://github.com/pangjunbiao/HO-LLN-Spoofing}.
\end{abstract}


\section{Introduction}
\label{sec:introduction}

Global Navigation Satellite System (GNSS) measurements, including GPS, provide an important source of absolute positioning information for autonomous vehicles~\cite{chen2023gps,zhang2025ghost}. However, GNSS signals are externally received and remain vulnerable to spoofing, in which a receiver is misled by counterfeit or manipulated positioning information~\cite{ying2023gps,yang2023location}. In autonomous driving, such
attacks can be safety-critical because an incorrect position estimate may propagate into localization, planning, and control decisions~\cite{shen2020drift,chen2023gps}. The problem is especially challenging when spoofing is continuous and subtle: the forged trajectory may
remain locally plausible while gradually becoming inconsistent with vehicle motion observed by independent onboard sensors. Effective detection therefore requires more than identifying an abrupt position jump; it requires causal modeling of weak GNSS--motion inconsistency as evidence that develops and persists over time.

Prior AV-oriented GNSS spoofing studies have shown that vehicle behavior provides useful evidence for detecting navigation attacks. Dasgupta \emph{et al.}~\cite{dasgupta2022sensor} proposed a sensor-fusion detector that compares an LSTM-predicted vehicle location shift with GNSS-derived motion and supplements it with a k-NN--DTW turn-detection branch. This demonstrates the value of GNSS--motion consistency, but its decision process relies mainly on
thresholded discrepancies and maneuver-specific checks rather than explicitly modeling the temporal development of weak inconsistency. Abrar \emph{et al.}~\cite{abrar2024gps} introduced GPS-IDS, which extracts physics-informed vehicle-behavior features from a GPS-integrated dynamic model and applies machine-learning classifiers to distinguish normal and spoofed navigation. Although this direction is valuable, the final detection stage is
largely formulated as feature-level anomaly classification. These studies therefore leave limited explicit treatment of how current residual disagreement, its local evolution, accumulated excess evidence, and observation validity should jointly support a confirmed spoofing decision under a causal information constraint.

To address this gap, this paper develops a causal sequential detector for continuous and subtle GNSS spoofing. First, the method compares the vehicle displacement indicated by consecutive GNSS positions with the displacement inferred from non-GNSS onboard motion observations. Their difference is normalized according to its expected uncertainty, so that the resulting evidence reflects the relative significance of the GNSS--motion disagreement rather than only its raw magnitude. The detector then represents the current inconsistency, its change over time, its excess above the normal level, and its persistence. These cues are organized into instantaneous, evolutionary, and persistent evidence roles. The three roles are aligned through a bounded symmetric interaction, and a third-order multiplicative interaction is then used to capture their joint activation. The resulting role-aware evidence is processed by a second-order liquid temporal model that maintains an evidence-memory state and an evidence-evolution state. Its response rates adapt to the current evidence, allowing weak but persistent inconsistencies to build over time while reducing the influence of isolated fluctuations. Finally, the detector estimates a spoofing probability at each time step, and an alarm is confirmed only after the probability exceeds a validation-selected threshold for a required number of consecutive time steps.

The main methodological contributions are summarized as follows:
\begin{itemize}
\item A physics-guided causal evidence representation is proposed to compare GNSS displacement with vehicle movement inferred from non-GNSS onboard sensors. It describes the current disagreement, how it changes, whether it exceeds the normal level, how long it persists, and whether the observations are valid.

\item A structured high-order interaction mechanism is developed to combine instantaneous, evolutionary, and persistent evidence. A Kirchhoff-inspired symmetric exchange enables balanced information sharing among the three evidence roles, while a third-order interaction captures their joint support for spoofing.

\item A second-order liquid sequential detector is introduced to track the combined evidence over time through coupled memory and evolution states. It produces a causal spoofing probability at each time step for persistence-confirmed alarm generation.

\end{itemize}

\section{Related Work}
\label{sec:relatedwork}

\subsection{GNSS Spoofing Detection and Multi-Sensor Consistency}
\label{sec:rw_gnss_consistency}

GNSS spoofing detection has been studied at signal, receiver, navigation-solution, and multi-sensor consistency levels. Rado\v{s} \emph{et al.}~\cite{radovs2024recent} reviewed jamming and spoofing defenses and noted that many direct methods require receiver observables, specialized processing, or extra hardware. Schmidt \emph{et al.}~\cite{schmidt2020gps} separated authentic and spoofed code-phase components using sparse LASSO modeling of correlator outputs, but this still depends on signal-domain access and thresholded decisions.

To reduce reliance on receiver-level observables, several works exploit consistency between GNSS and independent motion or localization sources. Clements \emph{et al.}~\cite{clements2022carrier} used carrier-phase GNSS, low-cost IMU, and vehicle constraints through tightly coupled residual costs. Chang \emph{et al.}~\cite{chang2022analytic} analyzed GNSS/INS/LiDAR Kalman-filter behavior under spoofing, while Shen \emph{et al.}~\cite{shen2020drift} showed that production-grade multi-sensor fusion can still be misled. External references have also been used: Oligeri \emph{et al.}~\cite{oligeri2022gps} compared GPS with cellular/Wi-Fi information, Wang \emph{et al.}~\cite{wang2023infrastructure} used secured roadside infrastructure with Isolation Forest detection, and Liu and Papadimitratos~\cite{liu2024extending} extended RAIM using opportunistic infrastructure and onboard sensing with a Gaussian-mixture spoofing likelihood. Although effective, these methods often depend on estimator residuals, fusion settings, communication, infrastructure, or likelihood thresholds. This motivates modeling onboard GNSS--motion disagreement as uncertainty-aware temporal evidence for subtle spoofing detection.

\subsection{Vehicle-Behavior and Learning-Based Spoofing Detection for Autonomous Vehicles}
\label{sec:rw_vehicle_behavior_learning}

AV-oriented spoofing detection has increasingly used vehicle-behavior cues. Dasgupta \emph{et al.} developed detectors based on low-cost in-vehicle sensors, including LSTM-based location-shift prediction, sensor fusion with motion-state and k-NN--DTW turn-consistency checking, and reinforcement-learning-based turn-by-turn detection~\cite{dasgupta2020prediction,dasgupta2022sensor,dasgupta2022reinforcement}. Yang \emph{et al.}~\cite{yang2023anomaly} used learning from demonstration with transportation and vehicle-engineering knowledge, and Shabbir \emph{et al.}~\cite{shabbir2023securing} evaluated SVM, CNN, and other learning models in CARLA-based simulations. Abrar \emph{et al.}~\cite{abrar2024gps} introduced GPS-IDS, a physics-informed detector that extracts temporal vehicle-behavior features from a GPS-integrated dynamic model on the AV-GPS dataset. These studies establish the value of behavior-level detection, but they mainly rely on predicted-shift thresholds, maneuver-specific checks, task-specific policies, or feature-level classification, rather than structured causal modeling of weak GNSS--motion evidence.

\subsection{Sequential Weak-Evidence Modeling and Adaptive Temporal Detection}
\label{sec:rw_sequential_temporal}

Continuous spoofing is sequential because a single residual may be ambiguous, whereas repeated weak inconsistencies can become reliable evidence. Time-series anomaly-detection reviews emphasize temporal context for contextual and collective deviations~\cite{blazquez2021review,schmidl2022anomaly}, and CUSUM-based detectors accumulate local statistics for temporary or non-stationary changes~\cite{watson2022sequential,liang2022quickest}. However, such methods usually require hand-designed statistics, distributional assumptions, and calibrated thresholds.

Deep temporal detectors learn sequence structure from multivariate data. OmniAnomaly uses a stochastic recurrent network~\cite{su2019robust}, USAD uses adversarial autoencoders~\cite{audibert2020usad}, MTAD-GAT and GDN model temporal or inter-sensor dependencies~\cite{zhao2020multivariate,deng2021graph}, and transformer-based detectors use attention-based anomaly scores~\cite{xu2021anomaly,tuli2022tranad}. Adaptive models such as liquid time-constant networks~\cite{hasani2021liquid} and neural controlled differential equations~\cite{kidger2020neural} further support input-dependent or continuous-time dynamics. Nevertheless, these methods mainly produce generic reconstruction, forecasting, dependency, or attention scores, without defining spoofing-specific evidence roles, high-order evidence interaction, or persistence-confirmed causal alarms. Therefore, existing work does not fully provide the required combination of uncertainty-normalized GNSS--motion residuals, causal weak-evidence representation, structured high-order interaction, adaptive liquid dynamics, and persistence-confirmed causal alarms.

\section{Methodology}

\subsection{Problem Formulation and Causal Detection Setting}
\label{sec:problem_formulation}

Consider an autonomous vehicle observed along a discrete trajectory $t=1,\ldots,T$. At each time step, the vehicle receives a GNSS observation $g_t\in\mathbb{R}^{d_g}$ and an onboard motion observation $u_t\in\mathbb{R}^{d_u}$. The GNSS observation contains the measured geographic position, whereas $u_t$ contains non-GNSS motion measurements used to infer the vehicle's movement. For $t\geq2$, let $\delta_t>0$ denote the elapsed time between observations at time steps $t-1$ and $t$.

The GNSS position channel is considered potentially compromised, while the onboard motion measurements used to construct the physical evidence are assumed not to be directly manipulated by the GNSS spoofer. Coordinated compromise of both observation channels is outside the scope of this work.

This work focuses on the causal detection of continuous and subtle GNSS spoofing. Unlike an abrupt attack that may create an immediately visible position discontinuity, subtle spoofing can preserve a locally plausible GNSS trajectory while gradually introducing disagreement between the displacement implied by GNSS positions and the vehicle motion inferred from onboard observations. The detection task is therefore formulated as a causal sequential problem in which weak GNSS--motion inconsistency must be assessed using only information available up to the current time.

Let $y_t\in\{0,1\}$ denote the ground-truth navigation state at time $t$, where $y_t=0$ represents normal GNSS operation and $y_t=1$ represents spoofing. The causal observation history available to the detector at time $t$ is defined as
\begin{equation}
\mathcal{H}_t
=
\left\{
\left(g_\tau,u_\tau,\delta_\tau\right)
\right\}_{\tau=1}^{t},
\label{eq:history}
\end{equation}
where $\delta_1=0$ is used for notational completeness. Given only $\mathcal{H}_t$, the detector estimates the time-step spoofing probability as
\begin{equation}
\hat{p}_t
=
F_{\Theta}\!\left(\mathcal{H}_t\right),
\qquad
\hat{p}_t\in[0,1],
\label{eq:general_detector}
\end{equation}
where $F_{\Theta}(\cdot)$ denotes the proposed causal sequential detector with learnable parameters $\Theta$. Since $\mathcal{H}_t$ contains no future observations, $\hat{p}_t$ satisfies the causal information constraint.

Each trajectory is processed independently and in chronological order. The temporal states of the detector are initialized at the beginning of the trajectory and then propagated using only current and past observations. During inference, the resulting time-step probabilities are converted into persistence-confirmed decisions using operating parameters selected only on the validation split.

\subsection{Physics-Guided GNSS--Motion Residual Evidence}
\label{sec:residual_evidence}

To instantiate the causal GNSS--motion inconsistency formulated in Section~\ref{sec:problem_formulation}, the detector compares the planar displacement implied by consecutive GNSS positions with the displacement inferred from onboard motion observations over the same time interval.

Let $\ell_t^{g}=[\mathrm{lat}_t,\mathrm{lon}_t]^{\top}$ denote the geographic position extracted from $g_t$. It is converted into a local east--north metric frame as
\begin{equation}
p_t^{g}
=
\mathcal{T}_{\mathrm{loc}}
\left(
\ell_t^{g};\ell_{\mathrm{ref}}
\right)
\in\mathbb{R}^{2},
\label{eq:local_gnss_position}
\end{equation}
where $\mathcal{T}_{\mathrm{loc}}(\cdot)$ denotes the local coordinate transformation and $\ell_{\mathrm{ref}}$ is a fixed reference coordinate.

For $t\geq2$, the GNSS-implied and onboard-motion-implied displacements over the interval $(t-1,t]$ are defined as
\begin{equation}
\Delta p_t^{g}
=
p_t^{g}-p_{t-1}^{g},
\qquad
\Delta p_t^{u}
=
\mathcal{K}
\left(
u_{t-1},u_t,\delta_t
\right),
\label{eq:displacement_pair}
\end{equation}
where $\mathcal{K}(\cdot)$ denotes a causal kinematic mapping that converts the available non-GNSS motion measurements into a planar displacement expressed in the same east--north frame as $\Delta p_t^{g}$. The mapping uses only measurements available up to time $t$.

A displacement pair is considered valid only when both displacement estimates are available and temporally aligned. Let $\nu_t\in\{0,1\}$ denote this validity indicator, where $\nu_t=1$ indicates a valid pair and $\nu_t=0$ otherwise. Since no consecutive displacement can be formed at the first observation, $\nu_1=0$. For a valid pair, the GNSS--motion displacement residual is
\begin{equation}
r_t
=
\Delta p_t^{g}-\Delta p_t^{u},
\qquad
r_t\in\mathbb{R}^{2}.
\label{eq:raw_residual}
\end{equation}

The raw residual can vary because of GNSS uncertainty, onboard motion-estimation error, sampling variation, and normal vehicle maneuvers. It is therefore normalized using a positive-definite residual uncertainty matrix $\Sigma_t\in\mathbb{R}^{2\times2}$, defined as

\begin{equation}
\Sigma_t
=
\begin{cases}
\Sigma_t^{\Delta g}
+
\Sigma_t^{\Delta u}
+
\epsilon_{\Sigma}I_2,
& \text{both covariances available},\\[1mm]
\widehat{\Sigma}_{r}^{\mathrm{tr}}
+
\epsilon_{\Sigma}I_2,
& \text{otherwise}.
\end{cases}
\label{eq:residual_uncertainty}
\end{equation}
Here, $\Sigma_t^{\Delta g}$ and $\Sigma_t^{\Delta u}$ denote the propagated covariances of the GNSS-implied and onboard-motion-implied displacements, respectively. In the first branch, the corresponding displacement-estimation errors are assumed uncorrelated. This branch is used only when reliable covariance estimates are available for both displacement estimates. Otherwise, $\widehat{\Sigma}_{r}^{\mathrm{tr}}$ is estimated from valid normal residuals in the training split and fixed thereafter. The regularization term $\epsilon_{\Sigma}I_2$, with $\epsilon_{\Sigma}>0$, ensures that $\Sigma_t$ is positive definite.

The uncertainty-normalized residual evidence is then defined as
\begin{equation}
\eta_t
=
\begin{cases}
\Sigma_t^{-\frac{1}{2}}r_t,
& \nu_t=1,\\[1mm]
\mathbf{0},
& \nu_t=0,
\end{cases}
\qquad
\eta_t\in\mathbb{R}^{2},
\label{eq:normalized_residual}
\end{equation}
where $\Sigma_t^{-1/2}$ denotes the symmetric inverse square root of $\Sigma_t$. For a valid displacement pair, $\eta_t^{\top}\eta_t=r_t^{\top}\Sigma_t^{-1}r_t$ is the corresponding uncertainty-normalized residual energy. The vector $\eta_t$ encodes the GNSS--motion disagreement after accounting for its expected scale and directional correlation, while $\nu_t$ distinguishes unavailable evidence from a valid pair with small residual disagreement. Together, $(\eta_t,\nu_t)$ provides the physical input to the subsequent modeling of evidence development and accumulation.

\subsection{Causal Evidence Development and Explicit Accumulation}
\label{sec:weak_evidence_representation}

The normalized residual $\eta_t$ describes the current GNSS--motion disagreement. However, an isolated residual may also arise from sensing noise, temporary motion-estimation error, or a short-duration maneuver. The detector therefore represents not only the current disagreement, but also how it changes and whether above-normal residual evidence persists over time. All quantities in this subsection are computed causally from current and past observations.

The local evolution of the residual evidence is described using causal finite-difference terms. For $t\geq2$, the residual change rate is defined as
\begin{equation}
\dot{\eta}_t
=
\begin{cases}
\dfrac{\eta_t-\eta_{t-1}}{\delta_t},
& \nu_t\nu_{t-1}=1,\\[2mm]
\mathbf{0},
& \text{otherwise},
\end{cases}
\label{eq:residual_rate}
\end{equation}
and, for $t\geq3$, its rate of change is
\begin{equation}
\ddot{\eta}_t
=
\begin{cases}
\dfrac{\dot{\eta}_t-\dot{\eta}_{t-1}}{\bar{\delta}_t},
& \nu_t\nu_{t-1}\nu_{t-2}=1,\\[2mm]
\mathbf{0},
& \text{otherwise},
\end{cases}
\label{eq:residual_acceleration}
\end{equation}
where $\bar{\delta}_t=(\delta_t+\delta_{t-1})/2$ accounts for the time separation between consecutive rate estimates. The initial values are set as $\dot{\eta}_1=\mathbf{0}$ and $\ddot{\eta}_1=\ddot{\eta}_2=\mathbf{0}$. These quantities are finite-difference descriptors of residual evolution rather than physical vehicle velocity or acceleration. They characterize whether the GNSS--motion disagreement remains stable, increases, decreases, or changes direction.

To determine whether the current disagreement exceeds its normal operating level, the baseline-compensated residual evidence is defined as
\begin{equation}
q_t
=
\begin{cases}
\eta_t^{\top}\eta_t-\mu_e-\kappa,
& \nu_t=1,\\[1mm]
0,
& \nu_t=0,
\end{cases}
\label{eq:baseline_compensated_evidence}
\end{equation}
where $\mu_e$ is the median uncertainty-normalized residual energy computed from valid normal samples in the training split, and $\kappa\geq0$ is a tolerance margin. Hence, $q_t>0$ indicates that the current residual energy exceeds the tolerated normal level $\mu_e+\kappa$, whereas $q_t\leq0$ provides no positive excess evidence.

The persistence of positive excess evidence is summarized through a nonnegative one-sided accumulation:
\begin{equation}
a_t
=
\begin{cases}
\max\!\left(0,\rho a_{t-1}+q_t\right),
& \nu_t=1,\\[1mm]
\rho a_{t-1},
& \nu_t=0,
\end{cases}
\qquad
a_0=0,
\label{eq:weak_evidence_accumulation}
\end{equation}
where $0<\rho\leq1$ controls evidence retention. Positive values of $q_t$ add excess evidence to the retained state, whereas negative values weaken or reset the accumulation. Sustained positive evidence therefore supports evidence build-up over time. When the residual is unavailable, no new evidence is added, and the existing accumulation is retained or gradually decayed through $\rho$. The parameters $\kappa$ and $\rho$ are selected using the validation split and then fixed for all subsequent evaluations.

To control its dynamic range, the accumulation is compressed as $\tilde{a}_t=\log(1+a_t)$. The compressed accumulation serves as an explicit persistence feature within $\xi_t$, while the learned temporal dynamics introduced later process the resulting evidence vectors causally over time.

The resulting causal evidence vector is
\begin{equation}
\xi_t
=
\left[
\eta_t^{\top},
\dot{\eta}_t^{\top},
\ddot{\eta}_t^{\top},
q_t,
\tilde{a}_t,
\nu_t
\right]^{\top}
\in\mathbb{R}^{9}.
\label{eq:causal_evidence_vector}
\end{equation}
The vector $\xi_t$ combines the current residual, its local evolution, above-normal residual evidence, accumulated persistence, and displacement validity. Including $\nu_t$ allows unavailable, zero-filled evidence to be distinguished from a valid observation with weak GNSS--motion disagreement.

\subsection{Conservative Role Alignment and Third-Order Evidence Interaction}
\label{sec:structured_high_order}

The causal evidence vector $\xi_t$ combines cues with different physical and temporal meanings. To preserve these distinctions before temporal modeling, its components are organized into three evidence roles: instantaneous, evolutionary, and persistent. Transient disturbances may appear strongly in
only one role, whereas sustained GNSS--motion inconsistency can produce coordinated activation across multiple roles. Each role is therefore mapped into a common latent space before their interaction is modeled.

The role-specific latent states are defined as
\begin{equation}
\begin{aligned}
s_t^{\mathrm{I}}
&=
\sigma\!\left(
W_{\mathrm{I}}
\left[\eta_t^{\top},q_t\right]^{\top}
+b_{\mathrm{I}}
\right),\\
s_t^{\mathrm{E}}
&=
\sigma\!\left(
W_{\mathrm{E}}
\left[
\dot{\eta}_t^{\top},
\ddot{\eta}_t^{\top}
\right]^{\top}
+b_{\mathrm{E}}
\right),\\
s_t^{\mathrm{P}}
&=
\sigma\!\left(
W_{\mathrm{P}}
\left[
\tilde{a}_t,\nu_t
\right]^{\top}
+b_{\mathrm{P}}
\right).
\end{aligned}
\label{eq:structured_evidence_states}
\end{equation}
where $\sigma(\cdot)$ is applied elementwise and $s_t^{i}\in(0,1)^{d_s}$ for $i\in\mathcal{G}=\{\mathrm{I},\mathrm{E},\mathrm{P}\}$. The instantaneous state represents the current residual and its excess above the normal level, the evolutionary state represents local residual change, and the persistent state represents accumulated evidence while retaining the current displacement-validity cue. Including $\nu_t$ allows the persistent role to distinguish current unavailability from a valid observation with weak residual evidence.

Before constructing the joint interaction, the three role states are conservatively aligned. For each pair $i,j\in\mathcal{G}$, $i\neq j$, a scalar alignment weight is computed as
\begin{equation}
C_{ij,t}
=
\sigma\!\left(
w_C^{\top}
\begin{bmatrix}
\left|s_t^{i}-s_t^{j}\right|\\
s_t^{i}\odot s_t^{j}
\end{bmatrix}
+b_C
\right),
\qquad
C_{ij,t}=C_{ji,t},
\label{eq:evidence_interaction_weight}
\end{equation}
where $\odot$ denotes elementwise multiplication. The absolute difference describes disagreement between two roles, while their elementwise product describes simultaneous activation. Because the descriptor is invariant to the ordering of $i$ and $j$ and the same parameters are shared across all pairs, the resulting alignment weights are symmetric.

Each role state is then updated as
\begin{equation}
\bar{s}_t^{i}
=
s_t^{i}
+
\frac{1}{2}
\sum_{\substack{j\in\mathcal{G}\\j\neq i}}
C_{ij,t}
\left(
s_t^{j}-s_t^{i}
\right),
\qquad
i\in\mathcal{G}.
\label{eq:structured_evidence_update}
\end{equation}
Since there are three roles and $C_{ij,t}\in(0,1)$, this update forms a convex combination of the current role state and the other two role states. It therefore keeps the aligned states bounded while allowing each role to incorporate information from the others. Moreover, symmetry causes the pairwise difference terms to cancel when summed across the three roles, yielding
$\sum_{i\in\mathcal{G}}\bar{s}_t^{i}
=
\sum_{i\in\mathcal{G}}s_t^{i}$.
This limited conservation property is Kirchhoff-inspired only in the sense that symmetric pairwise flows balance globally; no electrical-circuit model is assumed.

The aligned states are then combined through an explicit third-order interaction:
\begin{equation}
j_t
=
\bar{s}_t^{\mathrm{I}}
\odot
\bar{s}_t^{\mathrm{E}}
\odot
\bar{s}_t^{\mathrm{P}},
\qquad
j_t\in(0,1)^{d_s}.
\label{eq:third_order_evidence}
\end{equation}

The feature $j_t$ emphasizes latent dimensions in which instantaneous, evolutionary, and persistent evidence are jointly active. The aligned role states and their third-order interaction are finally fused as
\begin{equation}
\small
\zeta_t
=
\tanh\!\left(
W_H
\left[
\bar{s}_t^{\mathrm{I}};
\bar{s}_t^{\mathrm{E}};
\bar{s}_t^{\mathrm{P}};
j_t
\right]
+b_H
\right),
\quad
\zeta_t\in\mathbb{R}^{d_{\zeta}}.
\label{eq:fused_high_order_evidence}
\end{equation}
The resulting representation $\zeta_t$ integrates the three aligned evidence roles with their coordinated activation. Here, third-order refers specifically to the multiplicative interaction among the three aligned evidence-role states, rather than to the order of the temporal dynamics introduced subsequently.

\subsection{Second-Order Liquid Evidence Dynamics}
\label{sec:liquid_evidence_dynamics}

The role-aware representation $\zeta_t$ describes the coordinated evidence at the current time step. To model how this evidence is retained and changes over time, the detector maintains a learned evidence-memory state $h_t\in\mathbb{R}^{d_h}$ and an evidence-evolution state $v_t\in\mathbb{R}^{d_h}$. Whereas $a_t$ explicitly summarizes persistent excess residual energy, the state pair $(h_t,v_t)$ learns the temporal development of the complete role-aware evidence. Both states are initialized as $h_1=v_1=\mathbf{0}$ at the beginning of each trajectory.

For $t\geq2$, the current evidence representation first produces a candidate memory state:
\begin{equation}
\tilde{h}_t
=
\tanh\!\left(
W_c\zeta_t+b_c
\right),
\qquad
\tilde{h}_t\in(-1,1)^{d_h}.
\label{eq:candidate_evidence_state}
\end{equation}
The candidate $\tilde{h}_t$ represents the memory content suggested by the current evidence, while the final update also depends on the previously retained memory and its evolution.

To adapt the temporal response to the current evidence pattern, two positive time-constant vectors are estimated from $\zeta_t$:
\begin{equation}
\begin{aligned}
\tau_t^{h}
&=
\tau_{\min}^{h}\mathbf{1}
+
\left(\tau_{\max}^{h}-\tau_{\min}^{h}\right)
\sigma\!\left(W_{\tau h}\zeta_t+b_{\tau h}\right),\\
\tau_t^{v}
&=
\tau_{\min}^{v}\mathbf{1}
+
\left(\tau_{\max}^{v}-\tau_{\min}^{v}\right)
\sigma\!\left(W_{\tau v}\zeta_t+b_{\tau v}\right),
\end{aligned}
\label{eq:adaptive_time_constants}
\end{equation}
where $\mathbf{1}\in\mathbb{R}^{d_h}$ is the all-ones vector, $0<\tau_{\min}^{h}<\tau_{\max}^{h}$, and $0<\tau_{\min}^{v}<\tau_{\max}^{v}$. The corresponding memory-response and evolution-retention coefficients are
\begin{equation}
\gamma_t
=
\mathbf{1}
-
\exp\!\left(
-\frac{\delta_t}{\tau_t^{h}}
\right),
\qquad
\beta_t
=
\exp\!\left(
-\frac{\delta_t}{\tau_t^{v}}
\right),
\label{eq:liquid_response_coefficients}
\end{equation}
where division and exponentiation are applied elementwise. Thus, $\gamma_t,\beta_t\in(0,1)^{d_h}$ for $t\geq2$. A larger $\tau_t^{h}$ produces a slower memory response, whereas a larger $\tau_t^{v}$ retains the previous evolution direction for longer.

The coupled temporal states are updated as
\begin{equation}
\begin{aligned}
v_t
&=
\beta_t\odot v_{t-1}
+
\left(\mathbf{1}-\beta_t\right)
\odot
\left(\tilde{h}_t-h_{t-1}\right),\\
h_t
&=
h_{t-1}
+
\gamma_t\odot v_t,
\end{aligned}
\label{eq:second_order_liquid_update}
\end{equation}
where $\odot$ denotes elementwise multiplication. The first equation updates the evidence-evolution state by combining its previous direction with the discrepancy between the current candidate and the retained memory. The second equation updates the memory according to this evolution. Thus, current evidence is filtered through input-adaptive response and retention coefficients rather than being copied directly into the temporal state.

For $t\geq3$, the previous evolution state is determined by the change from $h_{t-2}$ to $h_{t-1}$. Eliminating this intermediate state therefore yields an equivalent memory-only recurrence involving both $h_{t-1}$ and $h_{t-2}$, which establishes the second-order temporal structure. The dynamics are liquid in the specific sense that their effective time constants, and hence their response rates, vary with the current evidence representation.

The time-step spoofing probability is obtained from the two temporal states:
\begin{equation}
\hat{p}_t
=
\sigma\!\left(
w_o^{\top}
\left[
h_t^{\top},
v_t^{\top}
\right]^{\top}
+b_o
\right),
\qquad
\hat{p}_t\in(0,1),
\label{eq:spoofing_probability}
\end{equation}
where $w_o\in\mathbb{R}^{2d_h}$ and $b_o\in\mathbb{R}$ are learnable output parameters.

\subsection{Training Objective and Persistence-Confirmed Inference}
\label{sec:training_and_inference}

The model parameters are learned using weighted binary cross-entropy over all labeled time steps in the training trajectories. For clarity, let $n$ index a training trajectory and let $\mathcal{I}_{\mathrm{tr}}$ denote the set of labeled trajectory--time pairs $(n,t)$. To ensure numerical stability, the predicted probabilities are clipped to
$[\epsilon_p,1-\epsilon_p]$, where $0<\epsilon_p<1/2$, and are denoted by $\hat{p}_{n,t}^{\,\epsilon}$. The training objective is
\begin{equation}
\begin{aligned}
\mathcal{L}_{\mathrm{det}}
=
-\frac{1}{|\mathcal{I}_{\mathrm{tr}}|}
\sum_{(n,t)\in\mathcal{I}_{\mathrm{tr}}}
\Big[
&\alpha_1 y_{n,t}
\log\!\left(\hat{p}_{n,t}^{\,\epsilon}\right)\\
&+
\alpha_0(1-y_{n,t})
\log\!\left(1-\hat{p}_{n,t}^{\,\epsilon}\right)
\Big],
\end{aligned}
\label{eq:weighted_bce_loss}
\end{equation}
where $\alpha_1$ and $\alpha_0$ compensate for class imbalance. The class weights are computed using only the training split and remain fixed thereafter.

During inference, each trajectory is processed independently. The time-step probability is converted into a binary decision as $c_t=\mathbb{I}(\hat{p}_t\geq\theta)$, where $c_t\in\{0,1\}$ and $\theta\in(0,1)$ is selected using the validation split. To suppress isolated probability excursions, a detection is confirmed only after $N_{\mathrm{p}}$ consecutive positive decisions:
\begin{equation}
\bar{c}_t
=
\begin{cases}
\displaystyle
\mathbb{I}\!\left(
\sum_{j=0}^{N_{\mathrm{p}}-1}c_{t-j}
=
N_{\mathrm{p}}
\right),
& t\geq N_{\mathrm{p}},\\[2mm]
0,
& t<N_{\mathrm{p}},
\end{cases}
\qquad
\bar{c}_t\in\{0,1\}.
\label{eq:confirmed_alarm}
\end{equation}
The threshold $\theta$ and confirmation length $N_{\mathrm{p}}\in\mathbb{N}_{+}$ are selected jointly from prespecified candidate sets using only the validation split and are then fixed for all subsequent evaluations. For each trajectory, the first time step satisfying $\bar{c}_t=1$ defines the confirmed detection time.

The framework represents persistence at three complementary levels. The scalar $a_t$ summarizes sustained excess physical residual energy, the coupled states $(h_t,v_t)$ learn the temporal development of the complete role-aware evidence, and $N_{\mathrm{p}}$ confirms persistence at the final decision level. These quantities serve distinct but complementary purposes rather than constituting redundant detection mechanisms. Since the state update, probability estimation, and alarm-confirmation rule use only current and past information, the complete detection process remains causal.

\begin{figure*}[!t]
\centering
\includegraphics[width=0.99\textwidth]{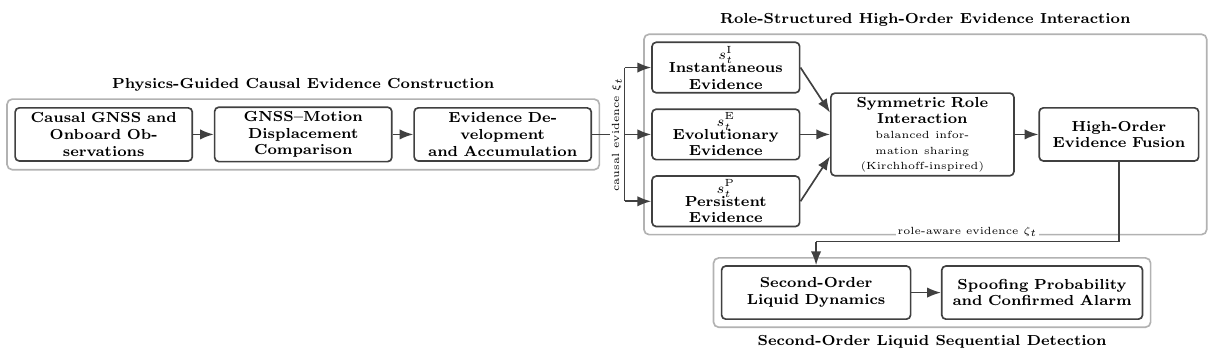}
\caption{Overview of the proposed causal high-order liquid evidence framework for continuous and subtle GNSS spoofing detection.}
\label{fig:method_overview}
\end{figure*}

\begin{table}[t]
\caption{Key notation used in the proposed framework.}
\label{tab:principal_notation}
\centering
\scriptsize
\renewcommand{\arraystretch}{1.08}
\setlength{\tabcolsep}{2.0pt}

\begin{tabularx}{\columnwidth}{
@{}
>{\centering\arraybackslash}p{2.25cm}
>{\raggedright\arraybackslash}X
@{}
}
\toprule
\textbf{Symbol}
& \textbf{Description} \\
\midrule

$g_t,\;u_t,\;\mathcal{H}_t$
& GNSS observation, non-GNSS onboard motion observation, and causal history up to time $t$ \\

$\Delta p_t^{g},\;\Delta p_t^{u}$
& GNSS-implied and onboard-motion-implied planar displacements \\

$r_t,\;\Sigma_t,\;\eta_t$
& GNSS--motion displacement residual, residual uncertainty matrix, and uncertainty-normalized residual evidence \\

$\nu_t$
& Displacement-validity indicator \\

$q_t,\;\tilde{a}_t$
& Above-normal residual evidence and compressed accumulated evidence \\

$\xi_t$
& Causal evidence vector combining the current residual, its evolution, persistence, and validity \\

$s_t^{\mathrm{I}},\;s_t^{\mathrm{E}},\;s_t^{\mathrm{P}}$
& Latent states representing instantaneous, evolutionary, and persistent evidence roles \\

$C_{ij,t}$
& Symmetric interaction weight between evidence roles $i$ and $j$ \\

$j_t$
& Third-order joint-activation feature of the three evidence roles \\

$\zeta_t$
& Fused role-aware high-order evidence representation \\

$h_t,\;v_t$
& Liquid evidence-memory and evidence-evolution states \\

$\tau_t^{h},\;\tau_t^{v}$
& Adaptive time constants controlling memory response and evolution retention \\

$\hat{p}_t,\;\bar{c}_t$
& Estimated spoofing probability and persistence-confirmation indicator \\

$\theta,\;N_{\mathrm{p}}$
& Validation-selected probability threshold and required number of consecutive positive decisions \\

\bottomrule
\end{tabularx}
\end{table}

\section{Experimental Results and Discussion}
\label{sec:results_discussion}

\subsection{Datasets, Evaluation Protocol, and Metrics}
\label{sec:datasets_protocol}

The experiments are conducted on the AV--GPS spoofing dataset collection~\cite{abrar2024gps}, which contains GNSS, vehicle-motion, and state variables recorded under normal and spoofed operation. The three AV--GPS scenarios are organized according to their experimental roles, as summarized in Table~\ref{tab:dataset_roles}. Dataset-1 is used for model training, validation-based operating-point selection, and held-out controlled comparison. Dataset-2 and Dataset-3 are reserved for evaluation of the finalized detector beyond the Dataset-1 test split, with Dataset-2 assessing external generalization and Dataset-3 serving as a causal sequential case study.

\begin{table}[H]
\centering
\caption{Dataset statistics and experimental roles of the AV--GPS scenarios.}
\label{tab:dataset_roles}
\scriptsize
\renewcommand{\arraystretch}{1.12}
\setlength{\tabcolsep}{2.0pt}
\begin{tabularx}{\columnwidth}{@{}lrrrr>{\raggedright\arraybackslash}X@{}}
\toprule
\textbf{Scenario} &
\textbf{Samples} &
\textbf{\shortstack{segments}} &
\textbf{Normal} &
\textbf{Spoofed} &
\textbf{Role in this paper} \\
\midrule
Dataset-1 & 62,042 & 27 & 46,287 & 15,755 & Model development and controlled comparison. \\
Dataset-2 & 6,890  & 6  & 5,184  & 1,706  & External generalization evaluation. \\
Dataset-3 & 636    & 1  & 231    & 405    & Causal sequential case study. \\
\bottomrule
\end{tabularx}
\end{table}

Dataset-1 is split at the trajectory-segment level so that samples from the same preprocessed trajectory segment do not appear in more than one partition. Its training partition is used to learn model parameters, its validation partition is used to select the operating point $(\theta,N_{\mathrm{p}})$, and its held-out test partition is used for controlled comparison. After this stage, the finalized detector, feature scaler, and operating point are applied unchanged to Dataset-2 and Dataset-3. To prevent leakage, all data-dependent statistics used for evidence construction and feature scaling are estimated only from the Dataset-1 training partition, including normal-residual statistics and the robust scaler for the continuous components of $\xi_t$. For fair comparison, all learning-based detectors use the same causal weak-evidence representation $\xi_t$, while shortcut-prone fields such as absolute GPS coordinates, time/date identifiers, and the EKF Detector output are excluded from the model inputs.

The evaluation reports threshold-independent ranking metrics, operating-point classification metrics, and event-level alarm metrics. The area under the receiver operating characteristic curve (AUROC) and the area under the precision--recall curve (AUPRC) are computed from the predicted spoofing probability $\hat{p}_t$ before thresholding. After applying the validation-selected threshold $\theta$ and persistence length $N_{\mathrm{p}}$, precision, recall, F1 score, and false-positive rate (FPR) are reported at the sample level. Event-level behavior is measured using the confirmed alarm sequence $\bar{c}_t$: attack detection rate (ADR) denotes the fraction of spoofing events with at least one confirmed alarm, and detection delay is the elapsed time from attack onset to the first confirmed alarm for detected events, with missed events reflected through ADR. For the Dataset-3 causal sequential case study, false-alarm rows and false-alarm events are also reported to quantify non-attack alarm samples and contiguous non-attack alarm episodes.

\subsection{Controlled Test and Cross-Scenario Evaluation}
\label{sec:controlled_external_results}

The evaluation proceeds in two stages. First, the held-out Dataset-1 test split is used to compare the proposed detector with representative learning-based alternatives under a controlled protocol. All learning-based detectors receive the same causal weak-evidence representation $\xi_t$, so the comparison reflects differences in the decision model rather than differences in input information. XGBoost and MLP are included as non-recurrent nonlinear classifiers, whereas LSTM and GRU are included as standard recurrent sequence models. Second, the finalized proposed detector is applied unchanged to Dataset-2 and Dataset-3 to examine its behavior beyond the Dataset-1 controlled setting.

Table~\ref{tab:dataset1_baseline_comparison} reports the Dataset-1 controlled test comparison. The proposed detector achieves the highest AUROC and AUPRC, together with the lowest false-positive rate among the learning-based detectors. XGBoost obtains the highest F1 score due to higher recall, but with a substantially longer detection delay. Although LSTM and GRU produce short delays, their false-positive rates are notably higher. Overall, the proposed detector provides the most favorable low-false-alarm operating trade-off while maintaining strong ranking performance and competitive classification accuracy.

\begin{table}[!t]
\centering
\caption{Dataset-1 controlled test comparison of learning-based detectors.}
\label{tab:dataset1_baseline_comparison}
\scriptsize
\renewcommand{\arraystretch}{1.10}
\setlength{\tabcolsep}{1.5pt}
\resizebox{\columnwidth}{!}{%
\begin{tabular}{@{}lccccccc@{}}
\toprule
\textbf{Model} &
\textbf{AUROC} $\uparrow$ &
\textbf{AUPRC} $\uparrow$ &
\textbf{F1} $\uparrow$ &
\textbf{Prec.} $\uparrow$ &
\textbf{Recall} $\uparrow$ &
\textbf{FPR} $\downarrow$ &
\textbf{Delay} $\downarrow$ \\
\midrule
Proposed & 0.9932 & 0.9843 & 0.9139 & 0.9909 & 0.8480 & 0.0027 & 3.5 \\
XGBoost & 0.9605 & 0.9514 & 0.9296 & 0.9835 & 0.8813 & 0.0052 & 14.5 \\
MLP & 0.9832 & 0.9650 & 0.8883 & 0.9667 & 0.8217 & 0.0099 & 8.5 \\
LSTM & 0.9466 & 0.9220 & 0.8704 & 0.9191 & 0.8266 & 0.0254 & 0.0 \\
GRU & 0.9627 & 0.9224 & 0.8467 & 0.9237 & 0.7814 & 0.0226 & 0.5 \\
\bottomrule
\end{tabular}%
}
\end{table}

Table~\ref{tab:finalized_detector_deployment} reports the cross-scenario performance of the finalized proposed detector. On Dataset-2, the detector maintains high AUROC, AUPRC, and strong classification performance, indicating that the learned evidence-to-decision mapping transfers beyond the Dataset-1 held-out test split. The ADR of 0.75 shows that one external spoofing event is not confirmed by the fixed persistence-based alarm rule, despite strong sample-level metrics. Dataset-3 is evaluated as a single causal sequential case study; the detector confirms both spoofing events and retains high AUPRC, but its false-positive rate and mean delay increase under continuous sequential operation. Overall, the results support sample-level generalization while revealing the stricter event-level trade-offs introduced by persistence-confirmed causal alarms.

\begin{table}[H]
\centering
\caption{Cross-scenario performance of the finalized proposed detector.}
\label{tab:finalized_detector_deployment}
\scriptsize
\renewcommand{\arraystretch}{1.10}
\setlength{\tabcolsep}{1.5pt}
\resizebox{\columnwidth}{!}{%
\begin{tabular}{@{}llcccccccc@{}}
\toprule
\textbf{Scenario} &
\textbf{Evaluation role} &
\textbf{AUROC} $\uparrow$ &
\textbf{AUPRC} $\uparrow$ &
\textbf{F1} $\uparrow$ &
\textbf{Prec.} $\uparrow$ &
\textbf{Recall} $\uparrow$ &
\textbf{FPR} $\downarrow$ &
\textbf{ADR} $\uparrow$ &
\textbf{Delay} $\downarrow$ \\
\midrule
Dataset-1 & Controlled held-out test & 0.9932 & 0.9843 & 0.9139 & 0.9909 & 0.8480 & 0.0027 & 1.00 & 3.5 \\
Dataset-2 & External generalization & 0.9899 & 0.9872 & 0.9567 & 0.9598 & 0.9536 & 0.0131 & 0.75 & 19.0 \\
Dataset-3 & Causal sequential case study & 0.9233 & 0.9661 & 0.8727 & 0.9205 & 0.8296 & 0.1266 & 1.00 & 35.0 \\
\bottomrule
\end{tabular}%
}
\end{table}

\subsection{Sequential Case Study on Dataset-3}
\label{sec:dataset3_case_study}

Dataset-3 is examined as a sequential case study to complement the aggregate results in Table~\ref{tab:finalized_detector_deployment}. Unlike the Dataset-1 controlled comparison, this scenario is processed as a single chronological sequence to inspect how the finalized detector behaves over time, including event-level alarm timing, non-attack alarm episodes, and the relationship between the probability output and the causal evidence traces.

\begin{figure}[!t]
\centering
\includegraphics[
width=\columnwidth,
height=0.40\textheight,
keepaspectratio]{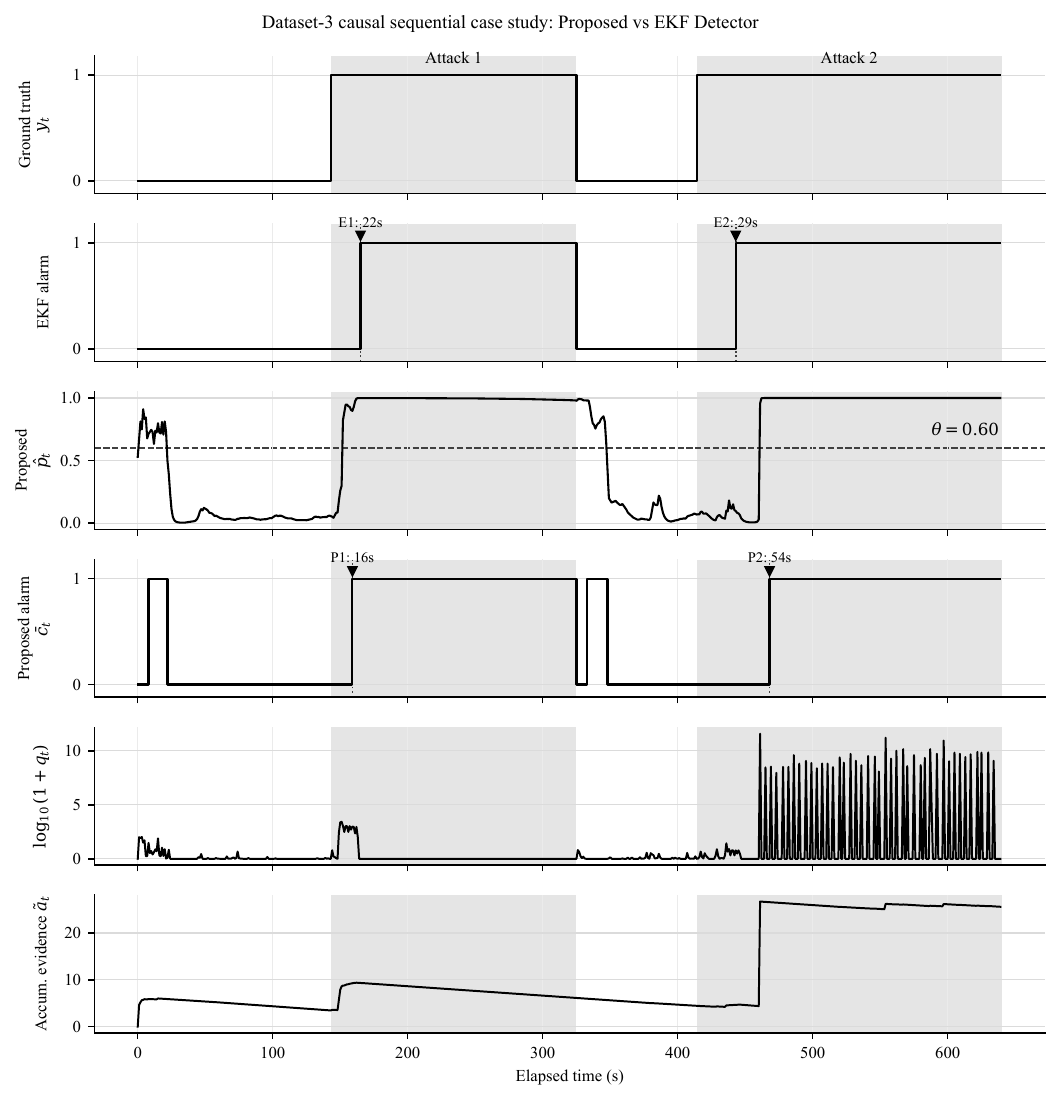}
\caption{Dataset-3 causal case study comparing the proposed detector and the EKF Detector during spoofing events.}
\label{fig:dataset3_case_study}
\end{figure}

Fig.~\ref{fig:dataset3_case_study} shows the sequential behavior of the proposed detector. The model produces a continuous spoofing probability $\hat{p}_t$, which is converted into the confirmed alarm sequence $\bar{c}_t$ after the validation-selected threshold $\theta$ and persistence length $N_{\mathrm{p}}$ are satisfied. During both spoofing events, the probability rises above the threshold and produces confirmed alarms. The evidence traces further show that these alarms are accompanied by the plotted residual-evidence trace $\log_{10}(1+q_t)$ and the accumulated residual evidence $\tilde{a}_t$, which is consistent with the subtle spoofing setting where evidence may emerge as persistent GNSS--motion inconsistency rather than a single abrupt position error. The traces also reveal several non-attack activations, explaining the higher false-positive rate observed on Dataset-3. Overall, the case study illustrates that the proposed detector provides a probability-based and evidence-traceable sequential response under continuous operation.

Table~\ref{tab:dataset3_event_level_comparison} reports the event-level comparison on the Dataset-3 sequence. Both methods detect the two spoofing events and achieve the same ADR. The proposed detector responds earlier to the first attack, whereas the EKF Detector responds earlier to the second attack, achieves a shorter mean delay, and produces no false-alarm episodes. Thus, the case study shows complementary behavior: the EKF Detector gives cleaner binary alarm timing on this sequence, while the proposed detector provides an interpretable probability-based response supported by its evidence trajectories.

\begin{table}[H]
\centering
\caption{Dataset-3 event-level comparison with the EKF Detector.}
\label{tab:dataset3_event_level_comparison}
\scriptsize
\renewcommand{\arraystretch}{1.10}
\setlength{\tabcolsep}{1.5pt}
\resizebox{\columnwidth}{!}{%
\begin{tabular}{@{}lcccccc@{}}
\toprule
\textbf{Method} &
\textbf{\shortstack{Attack-1\\Delay}} $\downarrow$ &
\textbf{\shortstack{Attack-2\\Delay}} $\downarrow$ &
\textbf{\shortstack{Mean\\Delay}} $\downarrow$ &
\textbf{ADR} $\uparrow$ &
\textbf{\shortstack{False\\Alarm Rows}} $\downarrow$ &
\textbf{\shortstack{False\\Alarm Events}} $\downarrow$ \\
\midrule
Proposed & 16.0 & 54.0 & 35.0 & 1.0 & 29 & 2 \\
EKF Detector & 22.0 & 29.0 & 25.5 & 1.0 & 0 & 0 \\
\bottomrule
\end{tabular}%
}
\end{table}

\subsection{Evidence-Role Analysis of the Causal Representation}
\label{sec:evidence_role_results}

The role of the causal weak-evidence representation is examined by ablating the evidence descriptors in $\xi_t$. The complete vector combines the instantaneous normalized residual $\eta_t$, residual-evolution descriptors $(\dot{\eta}_t,\ddot{\eta}_t)$, baseline-compensated residual evidence $q_t$, accumulated residual evidence $\tilde{a}_t$, and displacement validity $\nu_t$. We compare the complete representation with reduced settings that remove accumulated residual evidence, remove residual-evolution descriptors, remove both residual evolution and $q_t$, retain only low-order residual and validity information, or retain only the validity indicator. All settings follow the same training and evaluation protocol, so performance changes reflect the contribution of the available evidence roles rather than changes in experimental setup. This analysis tests whether subtle GNSS spoofing is better captured by combining instantaneous, evolutionary, persistent, and validity-aware evidence than by using instantaneous or validity cues alone.

Table~\ref{tab:evidence_component_d1} reports the Dataset-1 evidence-role analysis. The complete $\xi_t$ gives the most balanced behavior, combining strong ranking performance, high precision, low false-positive rate, complete attack detection, and short delay. Removing accumulated residual evidence $\tilde{a}_t$ keeps FPR low but sharply reduces recall, F1 score, ADR, and detection timeliness, showing the importance of persistence information for gradual spoofing. In contrast, removing residual-evolution descriptors, or removing both residual evolution and $q_t$, preserves high recall but greatly increases false positives, indicating that evolution and above-baseline residual evidence help suppress unstable alarms. The low-order and validity-only settings further confirm that residual magnitude or validity information alone is insufficient for reliable causal spoofing detection.

\begin{table}[!t]
\centering
\caption{Dataset-1 evidence-role analysis of the causal representation.}
\label{tab:evidence_component_d1}
\scriptsize
\renewcommand{\arraystretch}{1.10}
\setlength{\tabcolsep}{1.5pt}
\resizebox{\columnwidth}{!}{%
\begin{tabular}{@{}lcccccccc@{}}
\toprule
\textbf{Evidence-role setting} &
\textbf{AUROC} $\uparrow$ &
\textbf{AUPRC} $\uparrow$ &
\textbf{F1} $\uparrow$ &
\textbf{Prec.} $\uparrow$ &
\textbf{Recall} $\uparrow$ &
\textbf{FPR} $\downarrow$ &
\textbf{ADR} $\uparrow$ &
\textbf{Delay (s)} $\downarrow$ \\
\midrule
Complete evidence & 0.9932 & 0.9843 & 0.9139 & 0.9909 & 0.8480 & 0.0027 & 1.00 & 3.5 \\
Accumulation removed & 0.8814 & 0.7624 & 0.2137 & 0.9766 & 0.1200 & 0.0010 & 0.50 & 166.0 \\
Residual evolution removed & 0.9876 & 0.9765 & 0.6743 & 0.5126 & 0.9848 & 0.3274 & 1.00 & 3.5 \\
Evolution--$q_t$ removed & 0.9830 & 0.9723 & 0.6141 & 0.4462 & 0.9848 & 0.4274 & 1.00 & 3.5 \\
Low-order evidence only & 0.7284 & 0.4663 & 0.5154 & 0.3996 & 0.7256 & 0.3811 & 1.00 & 3.5 \\
Validity evidence only & 0.5018 & 0.2565 & 0.4105 & 0.2593 & 0.9840 & 0.9826 & 1.00 & 3.5 \\
\bottomrule
\end{tabular}%
}
\end{table}

Table~\ref{tab:evidence_component_cross_scenario} further shows that the evidence-role effects extend beyond Dataset-1. Removing accumulated residual evidence yields low FPR but produces an under-responsive detector, especially on Dataset-1 and Dataset-3. Conversely, removing residual-evolution descriptors, or removing both residual evolution and $q_t$, causes substantial false-positive inflation under external and sequential evaluation. The low-order and validity-only settings also show unstable cross-scenario behavior. These results support the design of $\xi_t$, which combines instantaneous, evolutionary, persistent, and validity-aware evidence for more stable causal detection of subtle GNSS spoofing.

\begin{table}[H]
\centering
\caption{Cross-scenario evidence-role analysis of the causal representation.}
\label{tab:evidence_component_cross_scenario}
\scriptsize
\renewcommand{\arraystretch}{1.10}
\setlength{\tabcolsep}{1.5pt}
\resizebox{\columnwidth}{!}{%
\begin{tabular}{@{}lcccccc@{}}
\toprule
\textbf{Evidence-role setting} &
\textbf{D1 F1} $\uparrow$ &
\textbf{D1 FPR} $\downarrow$ &
\textbf{D2 F1} $\uparrow$ &
\textbf{D2 FPR} $\downarrow$ &
\textbf{D3 F1} $\uparrow$ &
\textbf{D3 FPR} $\downarrow$ \\
\midrule
Complete evidence & 0.9139 & 0.0027 & 0.9567 & 0.0131 & 0.8727 & 0.1266 \\
Accumulation removed & 0.2137 & 0.0010 & 0.8977 & 0.0023 & 0.5081 & 0.0393 \\
Residual evolution removed & 0.6743 & 0.3274 & 0.5517 & 0.5031 & 0.7646 & 0.8210 \\
Evolution--$q_t$ removed & 0.6141 & 0.4274 & 0.4986 & 0.6226 & 0.7400 & 0.8559 \\
Low-order evidence only & 0.5154 & 0.3811 & 0.5515 & 0.4959 & 0.7785 & 0.6157 \\
Validity evidence only & 0.4105 & 0.9826 & 0.3927 & 0.9817 & 0.7910 & 0.9345 \\
\bottomrule
\end{tabular}%
}
\end{table}

\subsection{Architectural Role Analysis}
\label{sec:architecture_role_results}

After analyzing the causal weak-evidence representation, we examine the model-side architecture while keeping the full input $\xi_t$ fixed. This isolates the contribution of the structured decision model from the evidence representation itself. Starting from a recurrent-only variant, we progressively add the Kirchhoff-inspired symmetric exchange, the explicit third-order joint-activation feature, and the full liquid sequential detector. All variants use the same input representation and follow the same training and evaluation protocol.
\begin{table}[H]
\centering
\caption{Architectural role analysis using the common causal evidence input.}
\label{tab:high_order_liquid_analysis}
\scriptsize
\renewcommand{\arraystretch}{1.10}
\setlength{\tabcolsep}{1.5pt}
\resizebox{\columnwidth}{!}{%
\begin{tabular}{@{}llcccccc@{}}
\toprule
\textbf{Setting} &
\textbf{Model-side mechanism} &
\textbf{D1 F1} $\uparrow$ &
\textbf{D1 FPR} $\downarrow$ &
\textbf{D2 F1} $\uparrow$ &
\textbf{D2 FPR} $\downarrow$ &
\textbf{D3 F1} $\uparrow$ &
\textbf{D3 FPR} $\downarrow$ \\
\midrule
H0 & Recurrent-only variant & 0.9120 & 0.0474 & 0.7631 & 0.1809 & 0.7641 & 0.9083 \\
H1 & Kirchhoff-inspired symmetric exchange & 0.9345 & 0.0405 & 0.7136 & 0.2398 & 0.7796 & 1.0000 \\
H2 & Exchange + third-order feature & 0.9461 & 0.0167 & 0.9209 & 0.0454 & 0.8650 & 0.2140 \\
H3 & Full proposed & 0.9139 & 0.0027 & 0.9567 & 0.0131 & 0.8727 & 0.1266 \\
\bottomrule
\end{tabular}%
}
\end{table}

Table~\ref{tab:high_order_liquid_analysis} reports the architectural role analysis. The recurrent-only variant performs competitively on Dataset-1 but shows high false-positive rates under external and sequential evaluation, indicating limited cross-scenario stability. Adding only the symmetric exchange does not sufficiently reduce this instability, suggesting that pairwise evidence sharing alone is not enough. With the third-order feature, the F1--FPR balance improves substantially, especially on Dataset-2 and Dataset-3, supporting the value of joint activation among instantaneous, evolutionary, and persistent evidence roles. The full proposed architecture gives the most stable overall trade-off, achieving the lowest false-positive rate in all three scenarios and the highest F1 score on Dataset-2 and Dataset-3.

\subsection{Cross-Scenario Robustness Against Selected GPS-IDS Classifier Variants}
\label{subsec:gps_ids_comparison}

Beyond the common-input architectural evaluation, we assessed the cross-scenario robustness of the proposed detector against classifier variants of the GPS-IDS framework~\cite{abrar2024gps}. The proposed method operates on the corrected causal-evidence sequence, whereas the GPS-IDS branch uses a 15-variable representation reconstructed from the vehicle-dynamics, GPS-quality, navigation, and controller variables
specified in the original formulation.

All GPS-IDS variants followed the same experimental protocol and feature contract. All preprocessing parameters were estimated exclusively from the Dataset-1 training split. Model hyperparameters and alarm operating points were selected on the Dataset-1 validation split and subsequently frozen;
no samples from Dataset-1 test, Dataset-2, or Dataset-3 were used for fitting, tuning, or alarm calibration. To evaluate robustness across scenarios rather than performance weighted by dataset size,
Table~\ref{tab:gps_ids_cross_scenario} reports unweighted macro-averages over the three evaluation datasets, together with worst-case F1 and FPR.

\begin{table}[H]
\centering
\caption{Cross-Scenario Aggregate Comparison with Selected GPS-IDS Classifier Variants}
\label{tab:gps_ids_cross_scenario}
\scriptsize
\renewcommand{\arraystretch}{1.10}
\setlength{\tabcolsep}{1.5pt}
\resizebox{\columnwidth}{!}{%
\begin{tabular}{@{}lccccccc@{}}
\toprule
\textbf{Model} &
\textbf{Mean AUPRC} &
\textbf{Mean AUROC} &
\textbf{Mean F1} &
\textbf{Worst F1} &
\textbf{Mean FPR $\downarrow$} &
\textbf{Worst FPR $\downarrow$} &
\textbf{Mean ADR} \\
\midrule
\textbf{Proposed second-order liquid}
& \textbf{0.9792}
& \textbf{0.9688}
& \textbf{0.9144}
& \textbf{0.8727}
& \textbf{0.0475}
& \textbf{0.1266}
& \textbf{0.9167} \\

GPS-IDS--AdaBoost
& 0.9580
& 0.9335
& 0.7907
& 0.5630
& 0.3783
& 0.6407
& 0.7500 \\

GPS-IDS--MLP
& 0.7194
& 0.7248
& 0.7385
& 0.4426
& 0.5493
& 0.8230
& 0.8333 \\

GPS-IDS--Random Forest
& 0.6270
& 0.6899
& 0.7157
& 0.3970
& 0.6728
& 1.0000
& 0.6667 \\

GPS-IDS--Gradient Boosting
& 0.8406
& 0.8453
& 0.7091
& 0.3965
& 0.6763
& 0.9992
& 0.6667 \\

GPS-IDS--Decision Tree
& 0.5949
& 0.6265
& 0.6819
& 0.3831
& 0.6693
& 0.9985
& 0.8333 \\
\bottomrule
\end{tabular}%
}
\end{table}

Among the selected GPS-IDS variants, the proposed detector ranks first across all reported aggregate criteria. Its advantage is also retained in the worst-case measures, indicating that the overall result is not driven by a single favorable dataset. The concurrent improvements in discrimination, attack-event detection, and false-alarm control suggest that the proposed model capture gradually evolving spoofing evidence more reliably across different scenarios.

Because the two branches employ method-specific input representations, this experiment should be interpreted as an external-method robustness comparison rather than a strictly input-controlled architectural evaluation.

\subsection{Sensitivity and Multi-Seed Robustness}
\label{sec:sensitivity_multiseed}

We first conduct a one-factor sensitivity analysis by varying the probability threshold $\theta$, alarm persistence length $N_{\mathrm{p}}$, and hidden dimension while keeping the remaining settings fixed. We then assess training robustness by retraining the finalized model across five independent seeds and reporting the mean and standard deviation of each metric.

Table~\ref{tab:sensitivity_results} shows that the selected configuration lies in a favorable operating region. Lowering $\theta$ or reducing $N_{\mathrm{p}}$ slightly increases F1, but at the cost of higher false-positive rates. Increasing $\theta$ or $N_{\mathrm{p}}$ suppresses false alarms, but also increases delay and reduces F1. The hidden dimension shows a similar trade-off: the selected value of 64 gives the best F1--FPR balance, whereas a larger dimension increases the false-positive rate without improving F1.

\begin{table}[H]
\centering
\caption{Parameter sensitivity around the selected Dataset-1 configuration.}
\label{tab:sensitivity_results}
\footnotesize
\renewcommand{\arraystretch}{1.12}
\setlength{\tabcolsep}{5pt}
\begin{tabular}{@{}lcccc@{}}
\toprule
\textbf{Variant} &
\textbf{Value} &
\textbf{F1} $\uparrow$ &
\textbf{FPR} $\downarrow$ &
\textbf{Delay (s)} $\downarrow$ \\
\midrule

$\theta$ &
0.50 &
0.9231 &
0.0045 &
3.5 \\

$\theta$ &
\textbf{0.60} &
0.9139 &
0.0027 &
3.5 \\

$\theta$ &
0.90 &
0.8876 &
0.0000 &
15.5 \\

\midrule

$N_{\mathrm{p}}$ &
3 &
0.9231 &
0.0075 &
1.0 \\

$N_{\mathrm{p}}$ &
\textbf{8} &
0.9139 &
0.0027 &
3.5 \\

$N_{\mathrm{p}}$ &
11 &
0.9079 &
0.0013 &
5.5 \\

\midrule

Hidden dim. &
32 &
0.8978 &
0.0036 &
11.0 \\

Hidden dim. &
\textbf{64} &
0.9139 &
0.0027 &
3.5 \\

Hidden dim. &
128 &
0.9017 &
0.0260 &
4.5 \\

\bottomrule
\end{tabular}
\end{table}

Table~\ref{tab:multiseed_robustness} reports the mean and standard deviation across five training seeds. On Dataset-1, the detector shows small variation in AUROC, AUPRC, and F1, indicating that the controlled-test performance is not driven by a single random initialization. Dataset-2 also maintains stable AUROC and AUPRC under external generalization, while Dataset-3 retains stable AUPRC and F1 in the sequential case-study setting. The false-positive rate varies more on Dataset-2 and Dataset-3 than on Dataset-1, which is expected because external shift and continuous chronological operation make false-alarm behavior more sensitive.

\begin{table}[H]
\centering
\caption{Multi-seed robustness of the finalized proposed detector.}
\label{tab:multiseed_robustness}
\scriptsize
\renewcommand{\arraystretch}{1.12}
\setlength{\tabcolsep}{0pt}
\begin{tabular*}{\columnwidth}{@{\extracolsep{\fill}}lccc@{}}
\toprule
\textbf{Metric} &
\textbf{\shortstack{Dataset-1\\Test}} &
\textbf{\shortstack{Dataset-2\\External}} &
\textbf{\shortstack{Dataset-3\\Sequential}} \\
\midrule
Seeds &
5 &
5 &
5 \\
AUROC $\uparrow$ &
$0.9929{\pm}0.0025$ &
$0.9883{\pm}0.0019$ &
$0.9127{\pm}0.0194$ \\
AUPRC $\uparrow$ &
$0.9833{\pm}0.0059$ &
$0.9862{\pm}0.0018$ &
$0.9630{\pm}0.0065$ \\
F1 $\uparrow$ &
$0.9160{\pm}0.0073$ &
$0.9489{\pm}0.0315$ &
$0.8875{\pm}0.0142$ \\
Prec. $\uparrow$ &
$0.9790{\pm}0.0178$ &
$0.9375{\pm}0.0626$ &
$0.9373{\pm}0.0236$ \\
Recall $\uparrow$ &
$0.8607{\pm}0.0092$ &
$0.9625{\pm}0.0060$ &
$0.8430{\pm}0.0137$ \\
FPR $\downarrow$ &
$0.0066{\pm}0.0057$ &
$0.0224{\pm}0.0239$ &
$0.1004{\pm}0.0401$ \\
ADR $\uparrow$ &
$1.0000{\pm}0.0000$ &
$0.7500{\pm}0.0000$ &
$1.0000{\pm}0.0000$ \\
Delay (s) $\downarrow$ &
$5.0000{\pm}2.7839$ &
$16.7333{\pm}2.6604$ &
$32.3000{\pm}2.7749$ \\
\bottomrule
\end{tabular*}
\end{table}

\section{Conclusion}

This paper presented a high-order liquid evidence detector for continuous and subtle GNSS spoofing in autonomous driving. Rather than treating spoofing as a static anomaly, the proposed framework models the causal development of GNSS--motion inconsistency over time. It first derives uncertainty-normalized residual evidence by comparing GNSS-implied displacement with displacement inferred from independent onboard motion observations. The resulting causal evidence representation captures the current inconsistency, its local evolution, excess above the normal residual level, accumulated persistence, and displacement validity. To preserve their complementary physical and temporal roles, these cues are mapped into instantaneous, evolutionary, and persistent latent states, aligned through a bounded Kirchhoff-inspired symmetric exchange, and combined through an explicit third-order interaction. Second-order liquid dynamics with adaptive time constants then track the memory and evolution of the coordinated evidence to produce causal spoofing probabilities and persistence-confirmed alarms. Experiments on the AV--GPS dataset family demonstrate strong controlled and external generalization performance, together with clear sequential alarm behavior in the Dataset-3 causal case study. The proposed detector achieves high probability-ranking performance on Dataset-1 while maintaining the lowest false-positive rate among the learning-based baselines. The evidence-role and architectural analyses further indicate that residual evolution, accumulated persistence, coordinated high-order interaction, and liquid temporal modeling provide complementary contributions to detection. 

\subsection{Limitations and Future Work}

This study has three practical limitations. First, Dataset-3 contains only one chronologically ordered trajectory; it is therefore used as a causal sequential case study of detector behavior rather than as a broad deployment robustness benchmark. Second, residual-evidence construction depends on the temporal alignment of GNSS and independent onboard motion observations, and synchronization or motion-estimation errors may produce residual variations unrelated to spoofing. Third, the probability threshold and alarm persistence length are fixed after validation. Although this preserves the separation between validation and subsequent evaluation, recalibration may be required under substantial changes in vehicle platform, sensor configuration, driving environment, or spoofing pattern.

Future work will extend the sequential evaluation to additional continuous trajectories, heterogeneous sensor configurations, and more diverse spoofing patterns. It will also investigate adaptive operating-point calibration and richer multi-sensor evidence construction to improve alarm stability under changing deployment conditions.

\begingroup
\small 
\bibliographystyle{IEEEtran}
\bibliography{ref-2}

@article{abrar2024gps,
  title={GPS-IDS: An anomaly-based GPS spoofing attack detection framework for autonomous vehicles},
  author={Abrar, Murad Mehrab and Youssef, Amal and Islam, Raian and Satam, Shalaka and Latibari, Banafsheh Saber and Hariri, Salim and Shao, Sicong and Salehi, Soheil and Satam, Pratik},
  journal={arXiv preprint arXiv:2405.08359},
  year={2024}
}

@article{dasgupta2022sensor,
  title={A sensor fusion-based GNSS spoofing attack detection framework for autonomous vehicles},
  author={Dasgupta, Sagar and Rahman, Mizanur and Islam, Mhafuzul and Chowdhury, Mashrur},
  journal={IEEE Transactions on Intelligent Transportation Systems},
  volume={23},
  number={12},
  pages={23559--23572},
  year={2022},
  publisher={IEEE}
}

@inproceedings{clements2022carrier,
  title={Carrier-phase and IMU based GNSS spoofing detection for ground vehicles},
  author={Clements, Zachary and Yoder, James E and Humphreys, Todd E},
  booktitle={Proceedings of the ION International Technical Meeting, Long Beach, CA},
  pages={83--95},
  year={2022}
}

@inproceedings{shen2020drift,
  title={Drift with devil: Security of $\{$Multi-Sensor$\}$ fusion based localization in $\{$High-Level$\}$ autonomous driving under $\{$GPS$\}$ spoofing},
  author={Shen, Junjie and Won, Jun Yeon and Chen, Zeyuan and Chen, Qi Alfred},
  booktitle={29th USENIX security symposium (USENIX Security 20)},
  pages={931--948},
  year={2020}
}

@article{oligeri2022gps,
  title={GPS spoofing detection via crowd-sourced information for connected vehicles},
  author={Oligeri, Gabriele and Sciancalepore, Savio and Ibrahim, Omar Adel and Di Pietro, Roberto},
  journal={Computer Networks},
  volume={216},
  pages={109230},
  year={2022},
  publisher={Elsevier}
}

@article{wang2023infrastructure,
  title={Infrastructure-enabled GPS spoofing detection and correction},
  author={Wang, Feilong and Hong, Yuan and Ban, Xuegang},
  journal={IEEE transactions on intelligent transportation systems},
  volume={24},
  number={12},
  pages={13878--13892},
  year={2023},
  publisher={IEEE}
}

@article{radovs2024recent,
  title={Recent advances on jamming and spoofing detection in GNSS},
  author={Rado{\v{s}}, Katarina and Brki{\'c}, Marta and Begu{\v{s}}i{\'c}, Dinko},
  journal={Sensors},
  volume={24},
  number={13},
  pages={4210},
  year={2024},
  publisher={MDPI}
}

@article{schmidt2020gps,
  title={A GPS spoofing detection and classification correlator-based technique using the LASSO},
  author={Schmidt, Erick and Gatsis, Nikolaos and Akopian, David},
  journal={IEEE Transactions on Aerospace and Electronic Systems},
  volume={56},
  number={6},
  pages={4224--4237},
  year={2020},
  publisher={IEEE}
}

@article{chang2022analytic,
  title={Analytic models of a loosely coupled GNSS/INS/LiDAR Kalman filter considering update frequency under a spoofing attack},
  author={Chang, Jiachong and Zhang, Liang and Hsu, Li-Ta and Xu, Bing and Huang, Feng and Xu, Dingjie},
  journal={IEEE Sensors Journal},
  volume={22},
  number={23},
  pages={23341--23355},
  year={2022},
  publisher={IEEE}
}

@inproceedings{liu2024extending,
  title={Extending RAIM with a Gaussian mixture of opportunistic information},
  author={Liu, Wenjie and Papadimitratos, Panos},
  booktitle={Proceedings of the 2024 International Technical Meeting of The Institute of Navigation},
  pages={454--466},
  year={2024}
}

@misc{dasgupta2020prediction,
  author       = {Sagar Dasgupta and Mizanur Rahman and Mhafuzul Islam and Mashrur Chowdhury},
  title        = {Prediction-Based GNSS Spoofing Attack Detection for Autonomous Vehicles},
  year         = {2020},
  eprint       = {2010.11722},
  archivePrefix= {arXiv},
  primaryClass = {cs.CR}
}

@article{dasgupta2022reinforcement,
  author  = {Sagar Dasgupta and Tonmoy Ghosh and Mizanur Rahman},
  title   = {A Reinforcement Learning Approach for GNSS Spoofing Attack Detection of Autonomous Vehicles},
  journal = {Transportation Research Record},
  volume  = {2676},
  number  = {12},
  pages   = {318--330},
  year    = {2022},
  doi     = {10.1177/03611981221095509}
}

@article{yang2023anomaly,
  title={Anomaly detection against GPS spoofing attacks on connected and autonomous vehicles using learning from demonstration},
  author={Yang, Zhen and Ying, Jun and Shen, Junjie and Feng, Yiheng and Chen, Qi Alfred and Mao, Z Morley and Liu, Henry X},
  journal={IEEE Transactions on Intelligent Transportation Systems},
  volume={24},
  number={9},
  pages={9462--9475},
  year={2023},
  publisher={IEEE}
}

@article{shabbir2023securing,
  title={Securing autonomous vehicles against gps spoofing attacks: A deep learning approach},
  author={Shabbir, Maliha and Kamal, Mohsin and Ullah, Zahid and Khan, Maqsood Muhammad},
  journal={IEEE Access},
  volume={11},
  pages={105513--105526},
  year={2023},
  publisher={IEEE}
}

@article{blazquez2021review,
  title={A review on outlier/anomaly detection in time series data},
  author={Bl{\'a}zquez-Garc{\'\i}a, Ane and Conde, Angel and Mori, Usue and Lozano, Jose A},
  journal={ACM computing surveys (CSUR)},
  volume={54},
  number={3},
  pages={1--33},
  year={2021},
  publisher={ACM New York, NY, USA}
}

@article{schmidl2022anomaly,
  title={Anomaly detection in time series: a comprehensive evaluation},
  author={Schmidl, Sebastian and Wenig, Phillip and Papenbrock, Thorsten},
  year={2022}
}

@inproceedings{su2019robust,
  title={Robust anomaly detection for multivariate time series through stochastic recurrent neural network},
  author={Su, Ya and Zhao, Youjian and Niu, Chenhao and Liu, Rong and Sun, Wei and Pei, Dan},
  booktitle={Proceedings of the 25th ACM SIGKDD international conference on knowledge discovery \& data mining},
  pages={2828--2837},
  year={2019}
}

@inproceedings{audibert2020usad,
  title={Usad: Unsupervised anomaly detection on multivariate time series},
  author={Audibert, Julien and Michiardi, Pietro and Guyard, Fr{\'e}d{\'e}ric and Marti, S{\'e}bastien and Zuluaga, Maria A},
  booktitle={Proceedings of the 26th ACM SIGKDD international conference on knowledge discovery \& data mining},
  pages={3395--3404},
  year={2020}
}

@inproceedings{zhao2020multivariate,
  title={Multivariate time-series anomaly detection via graph attention network},
  author={Zhao, Hang and Wang, Yujing and Duan, Juanyong and Huang, Congrui and Cao, Defu and Tong, Yunhai and Xu, Bixiong and Bai, Jing and Tong, Jie and Zhang, Qi},
  booktitle={2020 IEEE international conference on data mining (ICDM)},
  pages={841--850},
  year={2020},
  organization={IEEE}
}

@inproceedings{deng2021graph,
  title={Graph neural network-based anomaly detection in multivariate time series},
  author={Deng, Ailin and Hooi, Bryan},
  booktitle={Proceedings of the AAAI conference on artificial intelligence},
  volume={35},
  number={5},
  pages={4027--4035},
  year={2021}
}

@article{xu2021anomaly,
  title={Anomaly transformer: Time series anomaly detection with association discrepancy},
  author={Xu, Jiehui and Wu, Haixu and Wang, Jianmin and Long, Mingsheng},
  journal={arXiv preprint arXiv:2110.02642},
  year={2021}
}

@article{tuli2022tranad,
  title={Tranad: Deep transformer networks for anomaly detection in multivariate time series data},
  author={Tuli, Shreshth and Casale, Giuliano and Jennings, Nicholas R},
  journal={arXiv preprint arXiv:2201.07284},
  year={2022}
}

@inproceedings{hasani2021liquid,
  title={Liquid time-constant networks},
  author={Hasani, Ramin and Lechner, Mathias and Amini, Alexander and Rus, Daniela and Grosu, Radu},
  booktitle={Proceedings of the AAAI conference on artificial intelligence},
  volume={35},
  number={9},
  pages={7657--7666},
  year={2021}
}

@article{kidger2020neural,
  title={Neural controlled differential equations for irregular time series},
  author={Kidger, Patrick and Morrill, James and Foster, James and Lyons, Terry},
  journal={Advances in neural information processing systems},
  volume={33},
  pages={6696--6707},
  year={2020}
}

@article{watson2022sequential,
  title={Sequential detection of a temporary change in multivariate time series},
  author={Watson, Victor and Septier, Fran{\c{c}}ois and Armand, Patrick and Duchenne, Christophe},
  journal={Digital Signal Processing},
  volume={127},
  pages={103545},
  year={2022},
  publisher={Elsevier}
}

@article{liang2022quickest,
  title={Quickest change detection with non-stationary post-change observations},
  author={Liang, Yuchen and Tartakovsky, Alexander G and Veeravalli, Venugopal V},
  journal={IEEE Transactions on Information Theory},
  volume={69},
  number={5},
  pages={3400--3414},
  year={2022},
  publisher={IEEE}
}

@article{chen2023gps,
  title={Gps attack detection and mitigation for safe autonomous driving using image and map based lateral direction localization},
  author={Chen, Qingming and Liu, Peng and Li, Guoqiang and Wang, Zhenpo},
  journal={arXiv preprint arXiv:2310.05407},
  year={2023}
}

@inproceedings{ying2023gps,
  title={Gps spoofing attack detection on intersection movement assist using one-class classification},
  author={Ying, Jun and Feng, Yiheng and Chen, Qi Alfred and Mao, Z},
  booktitle={ISOC Symposium on Vehicle Security and Privacy (VehicleSec)},
  year={2023}
}

@inproceedings{zhang2025ghost,
  title={The Ghost Navigator: Revisiting the Hidden Vulnerability of Localization in Autonomous Driving},
  author={Zhang, Junqi and Cheng, Shaoyin and Hu, Linqing and Zhang, Jie and Shi, Chengyu and Han, Xingshuo and Zhang, Tianwei and Cheng, Yueqiang and Zhang, Weiming},
  booktitle={34th USENIX Security Symposium (USENIX Security 25)},
  pages={3979--3998},
  year={2025}
}

@inproceedings{yang2023location,
  title={Location spoofing attacks on autonomous fleets},
  author={Yang, Jinghan and Estornell, Andew and Vorobeychik, Yevgeniy},
  booktitle={Symposium on Vehicles Security and Privacy},
  year={2023},
  organization={Internet Society}
}
\endgroup

\end{document}